# Institutional Foundations of Adaptive Planning: Exploration of Flood Planning in the Lower Rio Grande Valley, Texas, USA

Ashley D. Ross, [a]* Ali Nejat, [b] and Virgie Greb [a]

[a]*Department of Marine and Coastal Environmental Science, Texas A&M University at Galveston, Galveston, TX, USA;* [b]*Department of Civil, Environmental, and Construction Engineering, Texas Tech University, Lubbock, TX, USA*

*ashleydross@tamug.edu

## INTRODUCTION

Adaptive planning is ideally suited for the deep uncertainties presented by climate change. While there is a robust scholarship on the theory and methods of adaptive planning, this has largely neglected how adaptive planning is affected by existing planning institutions and how to move forward within the constraints of traditional planning organizations. This study asks: *How do existing traditional planning institutions support adaptive planning?* We explore this for flood planning in the Lower Rio Grande Valley of Texas, United States. We draw on county hazard plan and regional flood plan documents as well as transcripts of regional flood planning meetings to explore the emergent topics of these institutional outputs. Using Natural Language Processing to analyze this large amount of text, we find that hazard plans and discussions developing these plans are largely lacking an adaptive approach.

# INTRODUCTION

Planning for natural hazard risk reduction in the context climate change involves decision making under conditions of interacting, multiple uncertainties. Some of these are "deep uncertainties" connected to long time horizons, nonlinear changes in climates and ecosystems, and inability to reliably quantify the rate and magnitude of climate changes (Babovic & Mijic, 2018; Bosomworth & Gaillard, 2019). Other uncertainties are associated with the ambiguities and unpredictability of socioeconomic systems, including population growth, land use change, social conflict, and the whims of political will (Babovic & Mijic 2019; Buurman & Babovic, 2014). In the face of these uncertainties, a new paradigm of decision making has emerged that emphasizes the development of adaptive plans and policies (Hassnoot et al., 2013; Walker et al., 2013).

Traditional planning approaches typically generate a static optimal plan to reduce vulnerability to a single 'most likely' future or to respond a wide range of plausible future scenarios (Haasnoot et al., 2013; Manocha & Babovic, 2018). Because the future is largely unknowable, static optimal plans are likely to fail and adaptations are made ad-hoc to adjust to emerging risk conditions (Haasnoot et al., 2013). Decisions made are reactive and may create path dependency by locking in assets over long timeframes that entail high transfer costs when alternative options are pursued (Bloeman et al., 2018; Lawrence et al., 2019). In contrast, adaptive planning is dynamically robust. It accepts the multiple plausible futures resulting from deep uncertainties; commits to short-term actions that best respond to what is known; keeps long-term actions open; and adjusts actions dynamically in response to emergent risk conditions (Babovic & Mijic, 2018; Bosomworth & Gaillard 2019; Manocha & Babovic, 2018; Walker et al., 2013). Adaptive planning accepts that more information cannot resolve deep uncertainties; rather, the future must unfold to known what it holds (Buurman & Babovic, 2014). Key to adaptive planning is a framework of pre-specified contingency actions and associated monitoring mechanisms for triggers or tipping points over time that activate contingency actions if – and when – needed (Walker et al., 2013). As new information and environmental changes become evident, action then can be taken to adapt to revealed risk (Buurman & Babovic, 2014). Given the irreversibility, cost-intensity, and long lifespans of infrastructure, adaptive planning is preferable over traditional static planning approaches for guiding infrastructure investments (Manocha & Babovic, 2018). Moreover, studies of the application of adaptive planning have found that this approach increases decision maker awareness to uncertainties, provides political support for keeping long-term options open, and motivates decision makers to adjust plans in ways that better accommodate future conditions (Bloeman et al., 2018).

Despite the promise of adaptive planning, this approach is limited by inadequate consideration of existing institutions that may constrain decision processes and policy options available. Scholars have noted this tension, arguing:

> There exists a *re-adaptation challenge* in the scholarship and practice of adaptive governance that results from opposing tendencies and the widening gap between decision-making for future shocks (adaptation as it is widely understood) and the very real constraints imposed by entrenched institutional arrangements and physical infrastructure that is designed and operated for a narrowly defined set of conditions. (Scott et al., 2018, p. 104-105)

Natural hazard planning, in particular, is highly institutionalized as there are requirements and conditions set by federal governments for subnational entities to meet. In the United States (U.S.), disaster assistance and funding for hazard mitigation is made available by the Federal Emergency Management Agency (FEMA) only to those states and, by extension, localities that have approved hazard mitigation plans (Congressional Research Service, 2022). Not only do planning requirements present institutional constraints to adaptation, case studies have shown that adaptation options are circumscribed by decision cycles and choices of the past, made under specific institutional arrangements and in social and political contexts with asymmetric power (Eriksen & Lind, 2008; Tellman et al., 2018). Despite the reality that hazard planning takes place in an institutional environment with past legacies and current constraints, the scholarship on adaptative planning has not fully addressed this issue. As a result, the uptake and implementation of adaptive planning may be stalled or stymied.

To further the scholarship on adaptive planning, this study asks: *How do existing traditional planning institutions support adaptive planning?* We explore this for flood planning in the Lower Rio Grande Valley of Texas, U.S. We draw on county hazard plan and regional flood plan documents as well as transcripts of regional flood planning meetings to explore the emergent topics of these institutional outputs. Using Natural Language Processing to analyze this large amount of text, we find that hazard plans and discussions developing these plans are largely lacking an adaptive approach.

*Conceptual Framework of Adaptive Planning*

Adaptation refers to "a series of adjustments, measures or policies, to reduce the vulnerability or enhance the resilience of a system to observed or expected climate change, reducing damages and maximising potential opportunities" (Intergovernmental Panel on Climate Change [IPCC], 2007). Climate change is projected to increase global temperatures and sea level rise and cause more frequent and intense heat waves, droughts, hurricanes, tornadoes, and flooding (IPCC, 2018). Yet, climate projections are limited by "fundamental, irreducible uncertainties" arising "from limitations in knowledge (e.g., cloud physics), from randomness (e.g., due to the chaotic nature of the climate system), and from human actions (e.g., future greenhouse gas emissions)" (Dessai et al., 2009). In the context of flooding, these uncertainties are compounded by unknowns related to population growth, land use changes, and social behavior (Babovic & Mijic, 2019). Despite the challenges such uncertainties pose for planning, a number of adaptive planning approaches have emerged to manage hazard risk under uncertainty (Walker et al., 2013).

Adaptive Policymaking is a theoretical approach for designing dynamic robust plans in the presence of uncertainties (Ranger, et al., 2010; Walker et al., 2001). Adaptation over time is central to the planning process as "inevitable changes become part of a larger, recognized process and are not forced to be made repeatedly on an ad hoc basis" (Walker et al., 2013, p. 961). Using monitoring and corrective actions, long-term goals are maintained even as the knowledge base evolves. Analytical approaches for dealing with uncertainty have emerged with Adaptation Tipping Points and, by extension, Adaptation Pathways. Emerging from a need to respond to new climate scenarios, these approaches explore the sequencing a set of possible actions, given revealed information over time (Haasnoot et al., 2013). Rather than addressing specific scenarios, Adaptation Tipping Points and Adaptation Pathways focuses on the conditions under which a given plan of action will fail. Tipping points are conditions under which acceptable "technical, environmental, societal, or economic standards may be compromised" (i.e., acceptable level of flood risk). At the juncture of a tipping point, "the system cannot continue to perform as expected and requires the implementation of another adaptation action" (Manocha & Babovic 2018, p. 12). These prompt a 'transfer' from one pathway to another in the Adaptation Pathways approach, often envisioned as a metro map of different routes of action to take to get to desired targets or goals (Haasnoot et al., 2011). When tipping points occur is unknown; thus, when new information is revealed, only the timing of actions need to be updated – or a new pathway taken – to keep the plan from failing (Walker et al., 2013).

Combining these approaches, the Dynamic Adaptive Policy Pathways approach aims to build flexibility into a plan by sequencing the implementation of actions over time so that the system adapts to changing conditions (Haasnoot et al., 2013). Alternative sequences – pathways – are specified to deal with a range of plausible future conditions. Preferred pathways are established after weighing costs and benefits so that physically and socially robust actions form the core of the adaptive plan. How actions take advantage of or create new opportunities or co-benefits is also evaluated. The plan "covers short-term investment decisions, long-term options, and adaptation signals to identify when to implement actions or revisit decisions and consider alternate pathways" (Haasnoot, et al. 2020, p. 453).

All approaches to adaptive planning share three fundamental characteristics that we contend are critical for the transition from traditional hazard planning. One, adaptive planning approaches embrace uncertainty and offer a structured planning process to develop actions in the face of future ambiguity. Two, they rely on inclusive and diverse participation to frame the adaptation problem, select goals, identify means to achieve those goals, and evaluate outcomes of adaptation actions. Three, they facilitate the development of adaptation actions with co-benefits, which is particularly important in the face of climate extremes (i.e., drought and flood).

*Acceptance of Uncertainty*

All adaptive planning approaches rely on acceptance of uncertainty by decision makers (Haasnoot et al., 2013; Walker et al., 2013). In a review article on adaptive planning, Walker and colleagues emphasize how uncertainty is embedded in the process of adaptive planning, stating:

> The essential idea of planned adaptation is that planners facing deep uncertainty create a shared strategic vision of the future, explore possible adaptation strategies and pathways, commit to short-term actions, while keeping long-term actions open, and prepare a framework (including in some cases a monitoring system, triggers, and contingency actions) that guides future actions. Implicit in this is that planners accept the irreducible character of the uncertainties about the future and aim to reduce uncertainty about the expected performance of their plans. So, planners have to accept—and in a sense embrace—uncertainty, rather than spending large amounts of time and effort on trying to reduce it, and waiting to take action until the uncertainties have been resolved. (Walker et al., 2013, p. 970)

The challenges and constraints presented by deep uncertainties prompt decision makers to use the process of adaptive planning (Haasnoot et al., 2013; Lawrence & Haasnoot, 2017; Walker et al, 2013). Within this process, an acceptance of uncertainty frames shared visions and goals for the future. By extension, uncertainty is inherent to the adaptation strategies and pathways devised and supported by the public and decision makers. Adaptive planning seeks to reduce and manage uncertainty through the development of monitoring systems and corrective actions. Given that much of the uncertainty surrounding future climate conditions cannot be reduced (Dessai et al., 2009), adaptive planning also seeks out diverse participation from stakeholders to gather information from sectors and groups across the community, thereby reducing the uncertainty of information that is reducible.

*Diverse Participation*

Diverse stakeholder participation is critical in adaptive planning to adequately frame the adaptation problem, select goals, identify means to achieve those goals, and evaluate outcomes of adaptation actions. To achieve these ends, stakeholder participation must include the broad array of social sub-groups as well as public and private sectors in a community (Bosomworth & Gaillard, 2019). The inclusion of diverse groups and sectors sets the stage for knowledge sharing. Ambiguity about future hazards cannot be reduced by more information but rather by dialogue between stakeholders and decision-makers (Bosomworth & Gaillard, 2019). Funtowicz and Ravetz (2020) argue such complexity and uncertainty require a new methodology for science, one that "is not formalized deduction but an interactive dialogue" between all those who have a stake in the issue (p. 5). Elaborating on this, Rosa states:

> The essential function of quality assurance and critical assessment can no longer be performed only by a restricted corps of insiders (such as scientists and experts), the dialogue must be extended to all of those who have a stake in the issue, that is, to the extended peer community. When the problem situation is well defined (system uncertainties and decision stakes are both low) then normal science (positivist and reductionist science) will work well, but when the problems are poorly defined and there are great uncertainties potentially involving many actors and interests, then we must attend to this new production of knowledge. (Rosa, 2008, p. 24).

Beyond knowledge sharing, diverse participation in adaptive planning is needed ensure that adaptation decisions are equitable and truly address the drivers of risk (Eakin et al., 2021). Exposure and vulnerability to natural hazards is unevenly distributed globally and within populations, largely as a function of social (e.g., economic inequalities, discriminatory land use and zoning) rather than physical factors (Thomas et al., 2019). In the U.S., race and ethnic minority groups are disproportionately exposed to and affected by natural hazards (Fothergill et al., 1999; Tate et al., 2021). Compounding these inequities, existing power imbalances determine whose knowledge, values, and stakes are involved in planning (Bosomworth & Gaillard, 2019; Eriksen et al., 2015; Tellman et al., 2018). These power structures also determine what environmental signal or threat is responded to; when response matters and for whom; and what actions are taken, given the trade-offs of costs and benefits to dominant groups.

Structural inequities, however, are not overcome by simply creating a participatory space:

> Even presumably inclusive learning spaces are not immune from reproducing inequalities and exploiting inherent vulnerabilities. At the community level, as with the policy level, unequal power defines the range of developmental options conceivable and may close down trajectories that might address and overcome these inequities. (Tschakerta et al., 2016, p. 193)

Pursuing diverse and inclusive participation in planning, therefore, requires exploring if stakeholders have the capacity to fully participate (Bosomworth & Gaillard, 2019). Decision makers must seek out diverse perspectives and carefully consider procedural and distributive equity in terms of how people are able to participate and how the adaptation actions taken distribute resources and risk (Eakin et al., 2021). With the understanding gained through knowledge sharing as a result of participation from diverse social groups and sectors, adaptation actions may innovate to pursue actions that seek to reduce risk as well as provide other benefits (Ranger, et al., 2010).

*Development of Co-Benefits*

Adaptive planning is largely opportunistic in that it seeks to leverage developments in information, political will, and economic resources in making adaptations (Walker et al., 2013). Opportunities for "no-regret actions" with co-benefits to serve multiple societal goals (beyond climate adaptation and natural hazard risk reduction) are fundamental to climate adaptation (Berrang-Ford et al., 2011). Actions with multiple social benefits can legitimize or normalize otherwise potentially contentious investments (Runhaar et al., 2012) and, generally, improve social well-being and adaptive capacity for resilience (Dovers, 2009). The development of such interventions is best supported by intersectoral and diverse stakeholder participation throughout the planning process (Haasnoot et al., 2013; Walker et al., 2013). This participation may reveal pathways for the co-production and co-management of adaptation with the private sector (Tompkins and Eakin, 2012).

Adaptive planning to develop actions with co-benefits is particularly useful in the area of water control and resources. For example, the use of planning controls to prevent new developments in flood exposed areas would not only control flood risk but also benefit ecosystem restoration and enhance water quality through filtration (Ranger, et al., 2010). Scott and colleagues emphasize that adaptive planning offers mechanisms for water infrastructure planning that is *flexible*, a critical resilience property in the face of climate change. They state:

> As climate grows more unpredictable, the capacity to manage water in adaptive ways becomes a very valuable asset. Now, under climate change, we need to question how pre-existing water solutions can be re-adapted, if possible, to fit more adaptive governance approaches. In a climate change context, infrastructure can help in managing the timing, magnitude, and distribution of water flows by providing reservoirs, protections and barriers against water scarcity or excess. (Scott et al. 2020, p. 106)

Research has shown that adaptive planning can facilitate the design of interoperable water management systems that redirect water and make use of other systems to maintain or enhance performance function during exceedance events (O'Donnell et al., 2020). The ability of adaptive planning to facilitate the development of dynamically robust adaptive plans that feature actions with co-benefits or interoperable systems makes this approach uniquely suited for the water challenges climate change presents in terms of droughts, scarcity, and floods.

*Study Expectations*

Given their importance for adaptive planning, we expect the presence of a widespread acceptance of uncertainty, efforts to broaden diverse participation, and consideration of co-benefits of hazard mitigation actions in traditional planning to be evidence of institutions that may support adaptive planning. The successful transition from traditional to adaptive planning is beyond the scope of this study. Nonetheless, exploration of how fundamental tenets of adaptive planning – uncertainty, participation, and co-benefits – are present in traditional planning settings should advance our understanding of the way existing institutions may enable or constrain more adaptive thinking about natural hazards and climate change. We, therefore, should gain insights into the challenges of uptake and implementation of adaptive planning for public sector agencies that currently maintain traditional planning institutions.

## MATERIALS AND METHODS

*Study Area*

Located on the southernmost corner of Texas, the Lower Rio Grande Valley (LRGV) spans approximately 43,000 square miles, bordered by the Rio Grande River and Mexico to the west and south and the Gulf of Mexico to the east. It is comprised of four counties – Cameron, Hidalgo, Starr, and Willacy (see Figure 1). The region has a total population of 1.4 million persons of which 94% are Hispanic (Rio Grande Valley Connect [RGV], 2022).

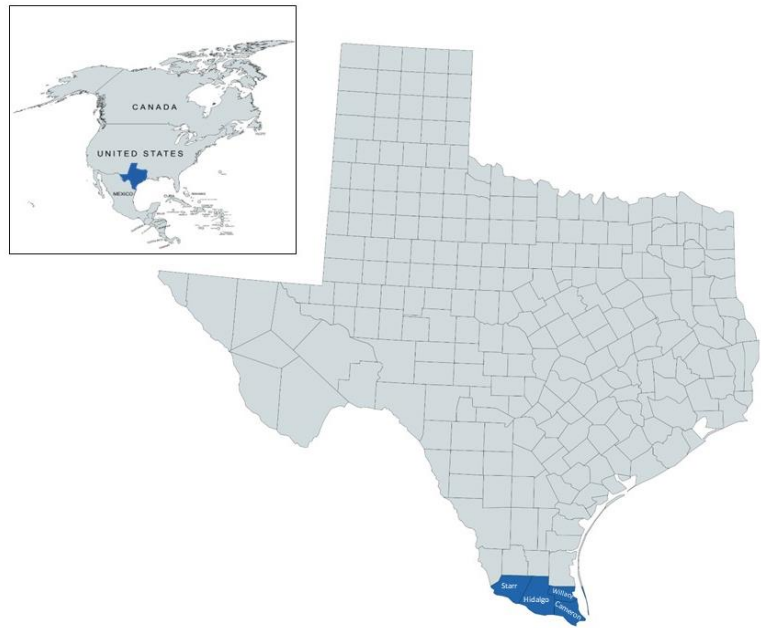

**Figure 1.** Study Area: Lower Rio Grande Valley, Texas, USA

Table 1 reports the population characteristics for each county in the study area. The LRGV is home to some of the largest – and fastest growing – cities in the nation. It is projected that the population of the Valley will nearly double by 2045 (to 2.4 million) due to significant employment growth (Rio Grande Valley Metropolitan Planning Organization, 2020). The main industries of the region include agriculture, manufacturing, oil and gas, transportation, and, most recently, space exploration with a SpaceX launch facility in Brownsville, Cameron County (RGV Connect, 2022; SpaceX Brownsville, 2022).

**Table 1.** County Population Characteristics

|  | CAMERON | HIDALGO | STARR | WILLACY |
|---|---|---|---|---|
| **GEOGRAPHY** | | | | |
| *Area (square miles)* | 892 | 1,571 | 1,223 | 591 |
| **POPULATION** | | | | |
| *Total* | 423,029 | 880,356 | 66,049 | 20,316 |
| *Under 18 years* | 29.3% | 31.6% | 32.5% | 23.4% |
| *65 years and older* | 14.0% | 11.3% | 11.3% | 14.0% |
| *Female* | 50.8% | 50.7% | 50.9% | 44.1% |
| *Hispanic* | 90.0% | 92.6% | 96.3% | 88.1% |
| **INCOME & POVERTY** | | | | |
| *Median household income* | $41,200 | $41,846 | $30,931 | $37,906 |
| *Poverty rate* | 24.4% | 23.9% | 25.2% | 24.7% |
| **EDUCATION** | | | | |
| *High school or higher* | 68.6% | 66.9% | 57.6% | 70.1% |
| *Bachelor's degree or higher* | 18.2% | 19.3% | 10.9% | 9.3% |

Source: U.S. Census Bureau

Since the early 1900s with the construction of railways and paved roads to connect the isolated region and irrigation canals to provide access to Rio Grande River water (Foscue, 1934), agriculture has been the dominant economic power in the LRGV. Employment is highly concentrated in agriculture work as well as personal and support services related to healthcare – all sectors with wages on the low-end of the scale (Rio South Texas Regional Cooperation Project, 2011). As a result of these employment trends and other socioeconomic legacies and patterns, socioeconomic vulnerability is high in the LRGV. Approximately 25% of families live below the Texas poverty line (RGV Connect, 2022). The average household income is $65,647, approximately $35,000 dollars below the average household income in Texas (RGV Connect, 2022). The counties in the study area are all within the top 10 counties in Texas for having a high percentage of poverty in the year 2020 (Texas Association of Counties, 2020). Additionally, there are approximately 900 *colonias* – small, unincorporated, rural communities – in the LRGV. These communities do not have clean drinking water, paved roads, or basic public services (Cantu, 2018). Socioeconomic vulnerability, therefore, and health issues are particularly acute in these communities.

Flooding has long been a major issue in LRGV due to its low-lying lands and proximity to the Gulf of Mexico. Severe storms and flooding in 2018 and 2019 impacted over 5,000 residences in the region and resulted in about $59 million in individual assistance provided by FEMA (FEMA, 2019; FEMA, 2018). Because of future climate variability, flooding events like these are more likely and will continue to present challenges. Moreover, continued urbanization and development poses flooding issues. The region has seen transformation from small, agricultural communities with acres of undeveloped land into suburban communities comprising the fifth-largest metropolitan area in Texas (EspeyConsultants, 2011). While urbanization and associated impervious cover have contributed to the flooding problem, the prevalence of older developments and unincorporated *colonias* further complicate it with insufficient infrastructure for drainage (EspeyConsultants, 2011; Uribe, n.d.).

At the same time, the LGRV has a history of seasons of drought and have begun to face water shortages due to urbanization and agriculture expansion (Norwine et al., 2004; Sanchez, 2022). The four-county region obtains a majority of its water supply from

Falcon-Amistad Lake/Reservoir System; currently Lake Falcon is only 9% full and Amistad is a 33% full (Baddour, 2022; Texas Water Development Board, n.d.a; b; c; d). Projections caution that water services to three million residents in South Texas will soon be severely constrained (Baddour, 2022). Heavy rainfall to recharge water sources and transporting water into the region are seen as the most viable solutions to current water shortages (Baddour, 2022).

*Flood Mitigation Planning in the Study Area*

The Disaster Mitigation Act of 2000 require that all local jurisdictions have an approved Hazard Mitigation Plan (HMP) in order to be eligible for any federal funding opportunities (Jackman & Beruvides, 2013). In the state of Texas, counties are required to prepare and adopt a HMP "with the primary purpose of identifying, assessing, and reducing the long-term risk to life and property from hazard events" (Texas General Land Office [TGLO], n.d.). Once county HMPs are approved, they must be updated every five years to maintain eligibility for various funding. Updating an HMP entails "reviewing and revising the plan to reflect changes in development, progress in local mitigation efforts, and changes in priorities" (TGLO, n.d.).

In addition to County Hazard Mitigation Plans (HMPs), the counties in the study area participate in regional planning efforts. In response to the impact of Hurricane Harvey and a number of devastating flood events across the state (Senate Research Center, 2019), the 86th Texas Legislature and Governor passed Senate Bill 8 on June 10, 2019 to create the first state flood plan. This legislation sets up a planning framework to support the development of updated flood risk products (e.g., floodplain maps, H&H models); enable regional participation and decision-making in the planning process; and encourage the development of interventions with co-benefits, particularly with regards to water exceedance and supply (Lake, 2021). Administered by the Texas Water Development Board (TWDB), the state flood plan relies on regional watershed planning (Lake, 2021). Fifteen regions, corresponding to major river and coastal basins as delineated by TWDB, have been designated across the state (Texas Living Waters, n.d.). Planning in these regions is coordinated by Regional Flood Planning Groups (RFPGs), comprised of 12 voting members, representing each of the following stakeholder groups: the public, agriculture, industry, river authorities, counties, municipalities, water districts, electric generating utilities, public water utilities, environmental interests, and small businesses (Texas Living Waters, n.d.).

In coordination with local stakeholders and technical advisors, RFPGs are tasked with assessing current flood mitigation strategies and policies as well as designating future projects based on these evaluations (Lake, 2021). According to Senate Bill 8:

> A regional flood plan must: 1) use information based on scientific data and updated mapping; and 2) include: a) a general description of the condition and functionality of flood control infrastructure in the flood planning region; b) flood control projects under construction or in the planning stage; c) information on land use changes and population growth in the flood planning region; d) an identification of the areas in the flood planning region that are prone to flood and flood control solutions for those areas; and e) an indication of whether a particular flood control solution meets an emergency need, uses federal money as a funding component, and may also serve as a water supply source. (Senate Bill 8, 2019)

Flood planning in the Lower Rio Grande Valley is coordinated by the Region 15 RFPG. Region 15 covers over 43,000 square miles across 14 counties, eight of which are only partially included in Region 15 (Region 15, 2022). The entirety of Cameron, Hidalgo, Starr, and Willacy Counties fall in the boundaries of Region 15.

To support regional flood mitigation, the 86th Texas Legislature also passed Senate Bill 7, designating $1.7 billion for flood control grants and loans (Anchondo, 2019). The bill created the Flood Infrastructure Fund (FIF), to be administered by TWDB, with a one-time appropriation of $793 million from the funds set aside by the bill (Lake, 2021). FIF awards provide grants and zero-percent interest loans to political jurisdictions for watershed planning; federal award matching funds; and structural (e.g., drainage systems, retention/detention infrastructure, levee networks) and non-structural (e.g., warning systems, public education) flood mitigation projects (TWDB 2020). To be eligible for funding, FIF projects must use best and most recently available data and information; coordinate with local stakeholders impacted by the project; and not duplicate other projects (Lake, 2021). Once the state flood plan is in place in September 2024, only projects identified in the plan will be eligible for FIF funding (Lake, 2021).

The first round of FIF funding has committed (as of May 31, 2022) $405 million and supported 126 active projects (TWDB, n.d.d). Priority was given to projects immediately protecting life and property and for watershed flood protection planning; to rural jurisdictions; and to jurisdictions with a flood-related federal disaster declaration in last five years (TWDB, 2020). FIF support totaling over $112 million has been committed to Cameron (totaling $45.9 million for 14 projects), Hidalgo (totaling $51.7 million for 6 projects), and Willacy (totaling $14.6 million for 14 projects) Counties while Starr County has not been awarded FIF support (TWDB, n.d.d). The FIF projects in the study counties are predominantly a combination of flood control planning, acquisition, design, construction, and rehabilitation. Additionally, watershed and flood protection studies are also being undertaken in Cameron and Willacy Counties.

*Data*

To explore how traditional flood planning institutions in the Lower Rio Grande Valley (LRGV) in the state of Texas may enable or constrain adaptive planning, we collected two sets of texts that may be understood as outputs of planning institutions and processes: *Corpus 1* – county and regional hazard and flood plans; and *Corpus 2* – regional flood planning group meeting transcripts. As content of formal hazard plans and planning discussions, we contend these data adequately provide insights into the degree to which adaptive thinking is present or not for flood planning in the LRGV.

Hazard and flood plans were collected for the study area from local government and regional planning group websites. Specifically, we include in the analysis Cameron County Hazard Mitigation Action Plan Update of 2021, totalling 470 pages; Hidalgo County Hazard Mitigation Action Plan Update 2021, totalling 1021 pages; and Starr County Multi-Jurisdictional Hazard Mitigation Plan for 2019-2024, totalling 98 pages. A Hazard Mitigation Plan was not included for Willacy County as it was not available online and requests made to local officials were not responded to. In addition to county hazard plans, the 2023 Regional Flood Draft Plan for Region 15 Lower Rio Grande Volume 1, totalling 223 pages, is included in the analysis. Volumes 2-4 of the regional plan were excluded as these include appendices with information beyond the scope of this analysis – maps, details on on-going mitigation projects, and facts sheets with cost evaluations for proposed projects.

In addition to the planning documents, meeting recordings from the Region 15 Flood Planning Group were downloaded from the group's website. Fifteen meetings, beginning with the group's first meeting on November 5, 2020 and ending with the meeting held on July 21, 2022, were professionally transcribed. Approximately 24 hours of meeting discussion was transcribed into text for analysis.

*Methods*

Natural Language Processing (NLP) is the branch of computer science which is mainly focused on the use of artificial intelligence to provide machines with the ability to understand text and linguistic data, virtually the same way that humans do (Campesato, 2022). The presence of a plethora of information has made the process of manual topic extraction almost impossible, leading to an increased demand in automating this process (Campos et al., 2020). Under the umbrella of NLP methodology, topic modeling is the process of extracting main keywords from vast bodies of linguistic data to identify major themes (Silge & Robinson, 2017).

Within the scholarship on natural hazards and disasters, the application of NLP and topic modeling has been used to explore effective and reliable disaster communication (Prabhakar Kaila, 2016), crisis communication requirements (Deng et al., 2020), compound disasters (Malakar & Lu, 2022), pandemic response (Cuaton & Neo, 2021), understanding public opining in various stages of disasters (Xu et al., 2019), identification of disaster risks (Gorro et al., 2021; Sakakibara et al., 2018), and post-disaster recovery needs (Jamali et al., 2019; Jamali et al, 2020). Only recently has NLP and topic modeling been used in research to explore planning documents. Brinkley and Stahmer (2021) pioneered the use of topic modeling in plan analysis to identify common areas of emphasis among Californian cities' general plans, finding more than 60 topics. In a similar study of 100 Resilient Cities plans, Fu and colleagues (2022) confirm that NLP techniques coincide with conventional content analysis and, thus, can be reliably used to explore plans, including those related to hazards and disasters. Further, the work by Lesnikowski and colleagues (2019) that applies topic modeling to the analysis of climate change policy speeches and meeting notes demonstrates the utility of NLP for data of that kind. Our study contributes to this line of research by exploring the text of hazard and flood plans (*Corpus* 1) as well as flood planning meeting discussion (*Corpus* 2) from the specific lens of adaptive planning.

For this study, we use used unsupervised keyword extraction for topic modeling purposes, which is less resource intensive than supervised methods. Under unsupervised keyword extraction methods, almost all algorithms follow the same sequence of tasks which includes pre-processing, candidate generation, candidate scoring, post-processing and ranking (Campesato, 2022). We adopt Latent Dirichlet allocation (LDA) to fit a topic model where each item of a text corpus is modeled as a finite mixture over an underlying set of topics through a three-level hierarchical Bayesian model (Blei et al., 2003). Simply put, LDA finds the topics of co-occurring words that maximize the probability of generating the original collection of documents. In modeling the topics most common in the corpus, LDA provides an Intertopic Distance Map that shows how similar the topics are in relation to one another. The size of the mapped topic circles indicates the marginal topic distribution or the proportion of words that belong to each topic across the corpus. Per NLP conventions, before running the LDA analysis we cleaned the text by removing stop words, punctuations, other unmeaningful words[1] and lemmatized to exclude similar words (Silge & Robinson, 2017).

---

[1] See Appendix A for list of words removed from the analysis.

**Results**

The objective of our analysis is to identify the topics of co-occurring words in two bodies of text that represent outputs of flood planning institutions in the Lower Rio Grande Valley: *Corpus 1* – county and regional hazard and flood plans and *Corpus 2* – regional flood planning group meeting transcripts. The results of the LDA analysis for *Corpus* 1 are provided in Figure 2. The Intertopic Distance Map is shown with five topics identified across two dimensions. The most salient and relevant words for each topic are listed in the table below the map. Salience refers to words that are most useful or informative for identifying a topic (Chuang et al., 2012) while relevance indicates how much a word belongs to a topic in comparison to all the other topics (Sievert & Shirley, 2014).

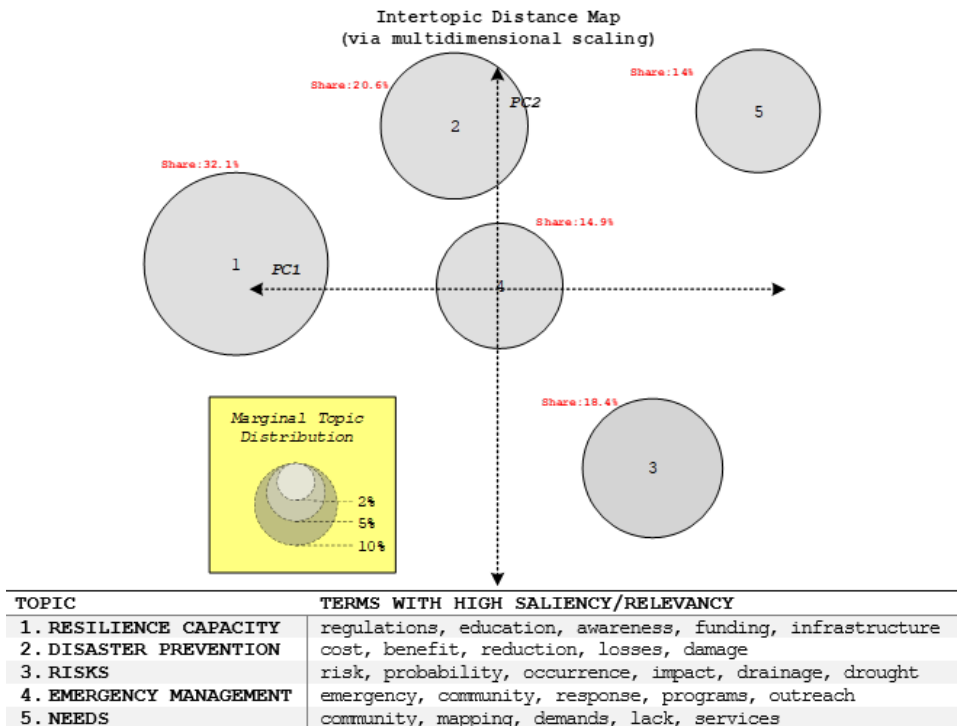

**Figure 2.** LDA Results for Corpus 1 – Hazard & Flood Plans

For Corpus 1, Topic 1, *resilience capacity*, spans a range of words that represent various community capacities to withstand, prevent, and manage a disaster event – 'regulations,' 'education,' 'awareness,' 'funding,' and 'infrastructure' (Ross, 2013). Topic 2, *disaster prevention*, includes words that indicate cost-benefit evaluations of actions to prevent or reduce losses and damages – 'cost,' 'benefit,' 'reduction,' 'losses,' and 'damage.' Topic 3, *risks*, encompasses mention of words that relate to the occurrence and severity of natural hazards – 'risk,' 'probability,' 'occurrence,' 'impact' – as well as words related to specific hazards in the study area – 'drainage' (related to flooding) and 'drought.' Topic 4, *emergency management*, covers words germane to this area of hazard management – 'emergency,' 'community,' 'response,' 'programs,' and 'outreach.' Topic 5, *needs*, includes words that we interpret as indicative of community needs for natural hazard risk reduction – 'community,' 'mapping,' 'demands,' 'lack,' and 'services.'

As shown by the size of the topic circles on the map for Corpus 1, Topics 1 and 2 – *resilience capacity* and *disaster prevention* – are the largest, meaning they include the largest share of words of the entire corpus. As shown by the placement of the topic circles on the map in relation to one another, Topics 1, 2, and 4 – *resilience capacity*, *disaster prevention*, and *emergency management* – have the most in common. Topics 3 and 5 –

*risks* and *needs* – are more isolated, indicated that they share fewer common words in relation to the other topics.

The results of the LDA analysis for *Corpus 2* are provided in Figure 3. Topic 1, *risk interpretation*, spans a range of words that represent the process of decision makers making sense of hazard risk in their planning discussions – 'risk,' 'floodplain,' 'conditions,' 'studies,' 'maps,' and 'guidance.' Topic 2, *decision making*, includes a wide range of words that taken together evoke the process of weighing costs and benefits for various public sector actions across the disaster cycle – 'recommendation,' 'activities,' 'mitigation,' 'preparedness,' 'response,' 'cost,' 'benefit,' 'fees,' 'government,' and 'jurisdictions.' Topic 3, *goals and strategies*, includes words that demonstrate discussion about plan vision and strategies to achieve it – 'goals,' 'strategies,' 'opportunities,' 'financing,' and 'retrofitting.' Topic 4, *planning process*, feature a set of words that we interpret as related to the process established by recent state legislation that leverages regional input and planning to develop a state-wide flood plan – 'legislative,' 'regulatory,' 'administrative,' 'feedback,' and 'changes.' Topic 5, *technical knowledge*, includes a set of words that refer to the interpretation and evaluation of technical knowledge provided by the consultant group to the regional flood planning board – 'meetings,' 'consultant,' 'understand,' 'fact,' and 'gaps.'

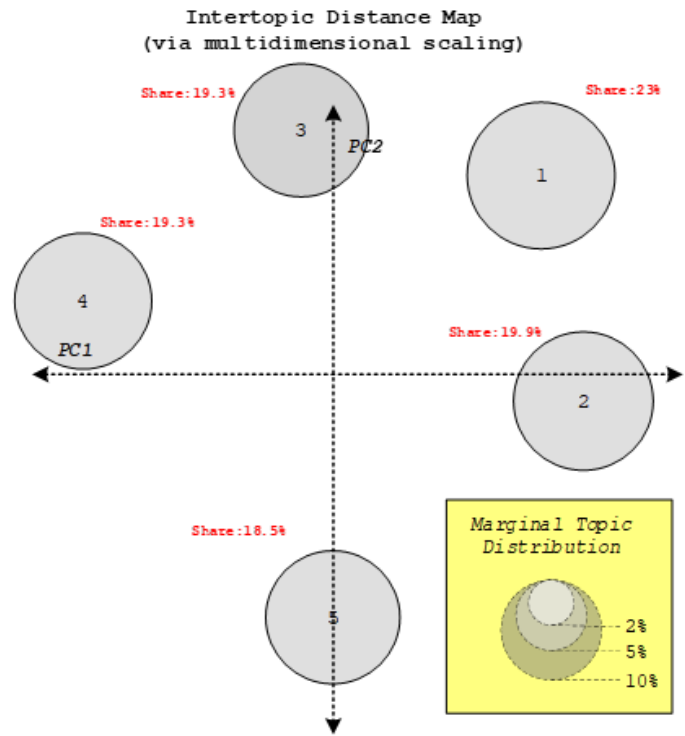

**Figure 3.** LDA Results for Corpus 2 – Flood Planning Meeting Transcripts

The topics for Corpus 2 are fairly balanced in terms of their share of words, spanning 19% (Topic 5 – *technical knowledge*) to 23% (Topic 1 – *risk interpretation*). In terms of commonality, Topics 1 and 2 – *risk interpretation* and *decision making* – are closely related whereas Topic 5 – *technical knowledge* – is the most isolated. Overall, the topics in Corpus 2 are mostly independent of one another.

**DISCUSSION**

The results of the analysis demonstrate that the central characteristics of adaptive planning – acceptance of uncertainty, inclusive public participation, and creation of co-benefits – are not the most dominant topics of the data analyzed for *Corpus 1* (county and regional hazard and flood plans) and *Corpus 2* (regional flood planning group meeting transcripts). Rather than approach hazards from a perspective of adaptation, there is a focus on *disaster prevention* (Topic 2 of Corpus 1), *emergency management* (Topic 4 of Corpus 1), and *decision making* around disaster mitigation, preparedness, and response (Topic 2 of Corpus 2). In fact, the word "adapt" is largely absent from the County HMPs and Region 15 Flood Plan. Only in the Cameron County HMP section on mitigation actions does the word "adaptive" appear in reference to a specific funded project to update the county-wide hydrologic and hydraulic (H&H) model. The Region 15 Flood Plan refers to "adapt" once when defining community resilience as "a measure of the ability of a community to prepare for anticipated natural hazards, adapt to changing conditions, and withstand and recover rapidly from disruptions" (Region 15, 2022, p. 2-36). Instead, the plans appear to focus on their mission to protect against loss of life and property from flood (and other) hazards. While this goal is fundamental for hazard planning, it does not evoke, or necessarily require, adaptation or adaptive thinking.

*Acceptance of Uncertainty*

The findings also show that rather than approach natural hazard planning with an acceptance of uncertainty, there is a focus on quantifying and interpreting risk (Topic 3 of Corpus 1 and Topic 1 of Corpus 2). Yet, the risk considered by the plans and meeting transcripts analyzed do not fully account for future conditions. The county HMPs rely on historic data and records to make inferences about the likelihood of future events. The Region 15 Flood Plan adopts current flood conditions to approximate future risk by utilizing the existing 0.2% annual chance event (ACE) or "500-year flood" area as a proxy for the future 1% ACE or "100-year" or "base" flood area. Further, in modeling flood hazards the plan acknowledges that it does not "consider projected changes in rainfall patterns, future land use/population growth, or planned new/improved infrastructure" (Region 15, 2022, p. 2-4). As described by Walker and colleagues, the fundamental downfall of a static approach like this is that it is likely to fail:

> Although policy analysts and strategic planners are aware that they are facing deep uncertainty, most of them still develop plans based on the assumption that the future can be predicted. They develop a static "optimal" plan using a single "most likely" future, often based on the extrapolation of trends, or a static "robust" plan that will produce acceptable outcomes in a small number of hypothesized future worlds. However, if the future turns out to be different from the hypothesized future(s), the plan is likely to fail. (Walker, Haasnoot, and Kwakkel, 2013, p. 957)

*Diverse Participation*

There is minimal emphasis on public participation in the results. The words 'education,' 'awareness,' and 'outreach' only occur as part of *resilience capacity* (Topic 1 of Corpus 1) and *emergency management* (Topic 4 of Corpus 1), rather than emerging as a topic on their own. Stakeholders involved in the development of the County HMPs included county and city representatives as well as irrigation and school districts, in some cases. Additionally, multiple public engagement efforts were part of the planning process including open public meetings to discuss hazards, the planning process, and solicit input on plans; survey instruments to capture preferences for mitigation action and problem areas identified by the public; and making the draft plan available for public review on participating jurisdictions' websites.

The Region 15 RFPG is led by multi-sector board, comprised of 14 voting members and 11 non-voting members. While the sectors included on the board were determined by the state and required of all RFPGs, it is notable that agriculture interests are represented by multiple voting and non-voting members in Region 15. Regarding public participation, the Region 15 RFPG has accepted written and public comments prior to and after all public meetings, of which there have been 14 since its inception in 2020 (*Region 15 Lower Rio Grande Regional Flood Planning Group*, n.d.). A survey was also distributed among community representatives and residents to "determine the nature of flood risk in regions, evaluate flood mitigation and management practices, and identify projects that reduce flood risk without negatively affecting neighboring areas" (*Lower Rio Grande Basin*, n.d.). Currently, public forums are planned in two locations in Region 15 to present and solicit public feedback on the draft flood plan.

While glimpses into the plans analyzed demonstrate there are multiple methods of public participation in the planning processes, it is not clear how inclusive these spaces were or what efforts were made to ensure access for all groups in the community. Diverse stakeholder participation is an integral part of the adaptive planning process because it serves as a mechanism of knowledge sharing required to fully address the social drivers of risk and design flexible and adaptive interventions to manage risk (Bosomworth & Gaillard, 2019). Such participation requires more than the creation of a space for public input (Tschakerta et al., 2016). It also requires acknowledgement that adaptation decisions inherently exclude "some risks and actors' interests while prioritizing others" (Eakin et al., 2021, p. 2); thus, structural inequities (e.g., power imbalances) and procedural justice (e.g., ability to participate) must be attended to ensure a pluralist process.

### *Development of Co-Benefits*

While there is an indication of potential adaptive thinking present in the topic of *goals and strategies* (Topic 3 of Corpus 2), no evidence of consideration of co-benefits is found. One of the most logical places for this to occur is in the Region 15 Flood Plan. The state required that all regional flood plans include an assessment to describe if any strategy or project: "1) involves directly increasing 'water supply' volume available during drought of record which requires both availability increase and directly connecting supply to specific water user group(s) with an identified water supply need; 2) directly benefits 'water availability' by, for example, injecting into aquifer but no one takes it as supply directly;  3) indirectly benefits 'water availability' (e.g., indirectly recharges aquifers naturally)" (TWDB, 2021). Region 15 determined that there were no anticipated impacts from their flood plan strategies or projects on water supply or availability. It appears that opportunities were not fully explored for multiple benefits of flood projects, which is unfortunate in an area routinely stricken by the climate extremes of flood and drought.

## CONCLUSION

This study has explored how traditional planning institutions may support adaptive planning. By doing so, we fill a gap in the scholarship on adaptive planning that largely neglects existing (and past) institutional legacies (Scott et al., 2018). We also add to the incipient body of research utilizing Natural Language Processing to analyze hazard plans (Brinkley & Stahmer, 2021; Fu et al., 2022). Further, and arguably most important, our study, by evaluating the process and inputs into the State of Texas' first state-wide flood plan, contributes to our understanding of a critical hazard plan as it develops.

The mandate by the Texas State Legislature in 2019 to develop a state-wide flood plan is a commendable institutional advancement towards adaptive planning. However, this study concludes that significant *re-adaptation challenges* remain (Scott et al., 2018). While the institutional foundations supporting flood planning in the study area are becoming more adaptive, our analysis shows limited technical capacity (e.g., lack of future flood risk data) and entrenched traditional approaches to hazard planning (e.g., neglect of uncertainty, lack of diverse public participation, disregard for adaptation co-benefits) among decision-makers may constrain how much progress is made toward adaptive planning.

This is particularly troubling because the state-wide plan relies on members of the regional planning groups to conduct analyses and designate flood mitigation projects that will be included in the state plan (Lake, 2021). Furthermore, only projects in the state plan are eligible for funding. Thus, there is high risk for path dependency. As a result, 'lock-in' situations could emerge that restrict the pool of possible options to maladaptive actions that are not well suited for unknown, future climate conditions (Bloeman et al., 2018; Haasnnot et al., 2019; Hetz & Bruns, 2014).

Scholarship on adaptive planning implies that to change this trajectory, transformative political action is needed (Bloeman et al., 2018; O'Donnell et al., 2020). We believe that such action must focus on enhancing the capacity of flood planning organizations and shifting organizational cultures from traditional (static optimal) to more adaptive (dynamic robust) thinking. Overcoming the present challenges in the study area cannot be achieved by institutional (legislative) change alone.

**Acknowledgement**
The authors would like to acknowledge Laura Button for her contribution to this paper through the National Science Foundation Ocean and Coastal Research Experiences for Undergraduates (OCEANUS) program at Texas A&M University at Galveston during the summer 2022.

**Data Availability**
Data is available upon request.

**Funding Details**
This work was supported by the National Science Foundation under a Smart and Connected Communities Planning Grant (SCC-PG), grant number NSF 21-535.

**Disclosure Statement**
The authors report there are no competing interests to declare.

**References**

Anchondo, Carlos. (2019, May 26). "Legislation with $1.7 billion for flood control and mitigation projects goes to governor." *Texas Tribune*. Accessed June 27, 2022 from: https://www.texastribune.org/2019/05/26/lawmakers-approve-bill-to-help-fund-floor-control-projects-in-texas/

*About Region 15*. (n.d.). Region 15 Lower Rio Grande Regional Flood Planning Group. Retrieved September 7, 2022, from http://www.region15lrg.org/page/homepage

Babovic, F., & Mijic, A. (2019). The development of adaptation pathways for the long-term planning of urban drainage systems. *Journal of Flood Risk Management*, *12*(S2), e12538.

Berrang-Ford, L., Ford, J. D., & Paterson, J. (2011). Are we adapting to climate change?. *Global environmental change*, *21*(1), 25-33.

Blei, D. M., Ng, A. Y., & Jordan, M. I. (2003). Latent dirichlet allocation. *Journal of machine Learning research*, *3*(Jan), 993-1022.

Bloemen, P., Reeder, T., Zevenbergen, C., Rijke, J., & Kingsborough, A. (2018). Lessons learned from applying adaptation pathways in flood risk management and challenges for the further development of this approach. *Mitigation and Adaptation Strategies for Global Change*, *23*(7), 1083-1108.

Bosomworth, K., & Gaillard, E. (2019). Engaging with uncertainty and ambiguity through participatory 'Adaptive Pathways' approaches: Scoping the literature. *Environmental Research Letters*, *14*(9), 093007.

Brinkley, C., & Stahmer, C. (2021). What is in a plan? using natural language processing to read 461 California city general plans. *Journal of Planning Education and Research*, 0739456X21995890.

Buurman, J., & Babovic, V. Design of Adaptive Policy Pathways under Deep Uncertainties An Approach to Climate Change Adaptation.

*Cameron County - Hazard Mitigation Action Plan Update 2021*. (2021, February 2). Cameron County. Retrieved September 8, 2022, from https://cameroncountytx.gov/wp-content/uploads/2021/02/2021-Cameron-County-HMP-s.pdf

*Cameron County, TX - Profile data*. (n.d.). Census Reporter. Retrieved September 8, 2022, from https://censusreporter.org/profiles/05000US48061-cameron-county-tx/

Campesato, O. (2022). *Natural Language Processing Using R Pocket Primer*. Mercury Learning and Information.

Campos, R., Mangaravite, V., Pasquali, A., Jorge, A., Nunes, C., & Jatowt, A. (2020). YAKE! Keyword extraction from single documents using multiple local features. *Information Sciences*, *509*, 257-289.

Chuang, J., Manning, C. D., & Heer, J. (2012, May). Termite: Visualization techniques for assessing textual topic models. In *Proceedings of the international working conference on advanced visual interfaces* (pp. 74-77).

*Colonias Prevention | Office of the Attorney General*. (n.d.). Texas Attorney General. Retrieved September 9, 2022, from https://www.texasattorneygeneral.gov/divisions/colonias-prevention

Congressional Research Service. (2022, July 27). Recent Funding Increases for FEMA Hazard
   Mitigation Assistance. IN11733. Version 10.

Cuaton, G. P., & Neo, J. F. V. (2021). A topic modeling analysis on the early phase of COVID-19 response in the Philippines. *International Journal of Disaster Risk Reduction*, *61*, 102367.

Deng, Q., Gao, Y., Wang, C., & Zhang, H. (2020). Detecting information requirements for crisis communication from social media data: An interactive topic modeling approach. *International Journal of Disaster Risk Reduction*, *50*, 101692.

Dessai, S., Hulme, M., Lempert, R., & Pielke Jr, R. (2009). Do we need better predictions to adapt to a changing climate?. *Eos, Transactions American Geophysical Union*, *90*(13), 111-112.

Dovers, S. (2009). Normalizing adaptation. *Global environmental change*, *19*(1), 4-6.

Eakin, H., Parajuli, J., Yogya, Y., Hernández, B., & Manheim, M. (2021). Entry points for addressing justice and politics in urban flood adaptation decision making. *Current Opinion in Environmental Sustainability*, *51*, 1-6.

*Economic Development - Demographics*. (n.d.). Cameron County. Retrieved September 7, 2022, from https://www.cameroncountytx.gov/economic-development-demographics/

Eriksen, S. H., Nightingale, A. J., & Eakin, H. (2015). Reframing adaptation: The political nature of climate change adaptation. *Global Environmental Change*, *35*, 523-533.

*Flood Planning Frequently Asked Questions (FAQ)*. (n.d.). Texas Water Development Board. Retrieved September 7, 2022, from https://www.twdb.texas.gov/flood/planning/faq.asp

Fothergill, A., Maestas, E. G., & Darlington, J. D. (1999). Race, ethnicity and disasters in the United States: A review of the literature. *Disasters, 23(2), 156-173.*

Funtowicz, S. O., & Ravetz, J. R. (1993). Science for the post-normal age. *Futures*, *25*(7), 739-755.

Fu, X., Li, C., & Zhai, W. (2022). Using Natural Language Processing to Read Plans: A Study of 78 Resilience Plans From the 100 Resilient Cities Network. *Journal of the American Planning Association*, 1-12.

Gorro, K. D., Baguia, G. A., & Ali, M. F. (2021, December). An analysis of Disaster Risk Suggestions using Latent Dirichlet Allocation and Hierarchical Dirichlet Process (Nonparametric LDA). In *2021 The 9th International Conference on Information Technology: IoT and Smart City* (pp. 181-184).

Haasnoot, M., Kwakkel, J. H., Walker, W. E., & Ter Maat, J. (2013). Dynamic adaptive policy pathways: A method for crafting robust decisions for a deeply uncertain world. *Global environmental change*, *23*(2), 485-498.

Haasnoot, M., van Aalst, M., Rozenberg, J., Dominique, K., Matthews, J., Bouwer, L. M., ... & Poff, N. L. (2020). Investments under non-stationarity: economic evaluation of adaptation pathways. *Climatic change*, *161*(3), 451-463.

Haasnoot, M.; Middelkoop, H.; van Beek, E.; van Deursen, W.P.A. A method to develop sustainable water management strategies for an uncertain future. Sust. Develop. 2011, 19, 369–381.

*Hazard Mitigation Assistance Grants*. (n.d.). FEMA. Retrieved September 8, 2022, from https://www.fema.gov/grants/mitigation

Hetz, K., & Bruns, A. (2014). Urban planning lock-in: implications for the realization of adaptive options towards climate change risks. *Water International*, *39*(6), 884-900.

*Hidalgo County Hazard Mitigation Action Plan - Update 2021*. (2021). www.hidalgocounty.us. Retrieved 09 08, 2022, from https://www.hidalgocounty.us/DocumentCenter/View/55458/Hidalgo-County-HMAP-APA-PUBLIC-COPY

*Hidalgo County, TX - Profile data*. (n.d.). Census Reporter. Retrieved September 8, 2022, from https://censusreporter.org/profiles/05000US48215-hidalgo-county-tx/

Intergovernmental Panel on Climate Change. (2007). Climate Change 2007: Synthesis Report. Contribution of Working Groups I, II and III to the Fourth Assessment Report of the Intergovernmental Panel on Climate Change. IPCC, Geneva, Switzerland.

Intergovernmental Panel on Climate Change. (2018). Global Warming of 1.5°C. An IPCC Special Report on the impacts of global warming of 1.5°C above pre-industrial levels and related global greenhouse gas emission pathways, in the context of strengthening the global response to the threat of climate change, sustainable development, and efforts to eradicate poverty. Cambridge University Press, Cambridge, UK.

Jackman, A. M., & Beruvides, M. G. (2013, 04 17). *Hazard Mitigation Planning in the United States: Historical Perspectives, Cultural Influences, and Current Challenges*. https://www.intechopen.com/chapters/44223. 10.5772/54209

Jamali, M., Nejat, A., Ghosh, S., Jin, F., & Cao, G. (2019). Social media data and post-disaster recovery. *International Journal of Information Management*, *44*, 25-37.

Jamali, M., Nejat, A., Moradi, S., Ghosh, S., Cao, G., & Jin, F. (2020). Social media data and housing recovery following extreme natural hazards. *International Journal of Disaster Risk Reduction*, *51*, 101788.

Lake, Peter. "Texas Reimagines the Fight Against Floods." *Texas Water Journal* 12, no. 1 (2021): 58-67.

Lawrence, J., & Haasnoot, M. (2017). What it took to catalyse uptake of dynamic adaptive pathways planning to address climate change uncertainty. *Environmental Science & Policy*, *68*, 47-57.

Lawrence, J., Allan, S., & Clarke, J. (2021). Using current legislative settings for managing the transition to a dynamic adaptive planning regime in New Zealand.

Lawrence, J., Bell, R., & Stroombergen, A. (2019). A hybrid process to address uncertainty and changing climate risk in coastal areas using dynamic adaptive pathways planning, multi-criteria decision analysis & real options analysis: a New Zealand application. *Sustainability*, *11*(2), 406.

Lawrence, J., Bell, R., Blackett, P., Stephens, S., Collins, D., Cradock-Henry, N., & Hardcastle, M. (2020). Supporting decision making through adaptive tools in a changing climate.

Lee, J., Kang, J. H., Jun, S., Lim, H., Jang, D., & Park, S. (2018). Ensemble modeling for sustainable technology transfer. *Sustainability*, *10*(7), 2278.

Lesnikowski, A., Belfer, E., Rodman, E., Smith, J., Biesbroek, R., Wilkerson, J. D., ... & Berrang-Ford, L. (2019). Frontiers in data analytics for adaptation research: Topic modeling. *Wiley Interdisciplinary Reviews: Climate Change*, *10*(3), e576.

Malakar, K., & Lu, C. (2022). Hydrometeorological disasters during COVID-19: Insights from topic modeling of global aid reports. *Science of The Total Environment*, 155977.

Manocha, N., & Babovic, V. (2017). Development and valuation of adaptation pathways for storm water management infrastructure. *Environmental Science & Policy*, *77*, 86-97.

Manocha, N., & Babovic, V. (2018). Real options, multi-objective optimization and the development of dynamically robust adaptive pathways. *Environmental science & policy*, *90*, 11-18.

O'Donnell, E., Thorne, C., Ahilan, S., Arthur, S., Birkinshaw, S., Butler, D., ... & Wright, N. (2020). The blue-green path to urban flood resilience. *Blue-Green Systems*, *2*(1), 28-45.

Prabhakar Kaila, D. (2016). An empirical text mining analysis of Fort McMurray wildfire disaster twitter communication using topic model. *Disaster Advances*, *9*(7).

Ranger, N., Millner, A., Dietz, S., Fankhauser, S., Lopez, A., & Ruta, G. (2010). Adaptation in the UK: a decision-making process. *Environment Agency*, *9*, 1-62.

Ranger, N., Millner, A., Dietz, S., Fankhauser, S., Lopez, A., & Ruta, G. (2010). Adaptation in the UK: a decision-making process. *Environment Agency*, *9*, 1-62.

Region 15 Flood Planning Group. (2022, July). Draft 2023 Flood Plan Volume 1. Retrieved September 1, 2022 from: http://www.region15lrg.org/upload/page/0071/docs/Region%2015%20RFP%20Draft_Volume%201.pdf

Rosa, M. P. (2008). Towards an adaptive approach in planning and management process. In *Integrated Water Management* (pp. 23-32). Springer, Dordrecht.

Ross, A. D. (2013). *Local disaster resilience: Administrative and political perspectives*. Routledge.

Runhaar, H., Mees, H., Wardekker, A., van der Sluijs, J., & Driessen, P. P. (2012). Adaptation to climate change-related risks in Dutch urban areas: stimuli and barriers. *Regional environmental change*, *12*(4), 777-790.

Sakakibara, H., Mori, S., Chosokabe, M., Kamiya, D., Yamanaka, R., Miyaguni, T., ... & Tsukai, M. (2018, October). The Topic Extraction from the Discussion Data of Community Disaster Risk Management Workshops. In *2018 IEEE International Conference on Systems, Man, and Cybernetics (SMC)* (pp. 462-467). IEEE.

Scott, C. A., Shrestha, P. P., & Lutz-Ley, A. N. (2020). The re-adaptation challenge: limits and opportunities of existing infrastructure and institutions in adaptive water governance. *Current Opinion in Environmental Sustainability*, *44*, 104-112.

Senate Bill 8, 86th Legislature, Regular Session. (Texas 2019). Accessed June 27, 2022 from: https://capitol.texas.gov/tlodocs/86R/billtext/pdf/SB00008F.pdf#navpanes=0

Senate Research Center, Bill Analysis, Senate Bill 8, 86th Legislature, Regular Session. (2019). Accessed June 27, 2022 from: https://capitol.texas.gov/tlodocs/86R/analysis/pdf/SB00008I.pdf#navpanes=0

Sievert, C., & Shirley, K. (2014, June). LDAvis: A method for visualizing and interpreting topics. In *Proceedings of the workshop on interactive language learning, visualization, and interfaces* (pp. 63-70).

Silge, J., & Robinson, D. (2017). *Text mining with R: A tidy approach*. O'Reilly Media, Inc.

*Starr County, TX - Geographic Facts & Maps*. (n.d.). MapSof.net. Retrieved September 7, 2022, from https://www.mapsof.net/starr-county

*Starr County, TX - Profile data*. (n.d.). Census Reporter. Retrieved September 8, 2022, from https://censusreporter.org/profiles/05000US48427-starr-county-tx/

Tate, E., Rahman, M. A., Emrich, C. T., & Sampson, C. C. (2021). Flood exposure and social vulnerability in the United States. *Natural Hazards, 106(1), 435-457*.

Tellman, B., Bausch, J., Eakin, H., Anderies, J., Mazari-Hiriart, M., Manuel-Navarrete, D., & Redman, C. (2018). Adaptive pathways and coupled infrastructure: seven centuries of adaptation to water risk and the production of vulnerability in Mexico City. *Ecology and Society*, *23*(1).

*Texas Counties by Population*. (n.d.a). Texas. Retrieved September 7, 2022, from https://www.texas-demographics.com/counties_by_population

Texas Living Waters (n.d.b) "Texas State and Regional Planning." Accessed June 27, 2022 from: https://texaslivingwaters.org/deeper-dive/rfpg_factsheet/

Texas Water Development Board. (n.d.c). *2023 Regional Flood Plan Region 15 Lower Rio Grande*.

Texas Water Development Board. (n.d.d). Flood Infrastructure Fund. Accessed June 27, 2022: https://www.twdb.texas.gov/financial/programs/fif/index.asp

Texas Water Development Board. (2020, September 17). 2020 Flood Intended Use Plan. Accessed June 27, 2022 from:

https://www.twdb.texas.gov/financial/programs/fif/doc/2020_Flood_Intended_Use_Plan.pdf?d=13515

Texas Water Development Board. (2021, April). Technical Guidelines for Regional Flood Planning. Accessed September 1, 2022 from: https://www.twdb.texas.gov/flood/planning/planningdocu/2023/doc/04_Exhibit_C_TechnicalGuidelines_April2021.pdf

*The Demographic Statistical Atlas of the United States*. (n.d.). The Demographic Statistical Atlas of the United States - Statistical Atlas. Retrieved September 7, 2022, from https://statisticalatlas.com/county/Texas/Hidalgo-County/Population

Thomas, K., Hardy, R. D., Lazrus, H., Mendez, M., Orlove, B., Rivera-Collazo, I., ... & Winthrop, R. (2019). Explaining differential vulnerability to climate change: A social science review. *Wiley Interdisciplinary Reviews: Climate Change, 10(2), e565.*

Tomas, R. C., Manuta, J. B., & dela Rosa, V. G. (2011). Too Much or Too Little Water: Adaptation Pathways of Agusan Marsh Communities. *SLONGAN*, *1*(1), 14-14.

Tompkins, E. L., & Eakin, H. (2012). Managing private and public adaptation to climate change. *Global environmental change*, *22*(1), 3-11.

Tschakert, P., Das, P. J., Pradhan, N. S., Machado, M., Lamadrid, A., Buragohain, M., & Hazarika, M. A. (2016). Micropolitics in collective learning spaces for adaptive decision making. *Global Environmental Change*, *40*, 182-194.

*U.S. Census Bureau QuickFacts: Hidalgo County, Texas*. (n.d.). U.S. Census Bureau. Retrieved September 7, 2022, from https://www.census.gov/quickfacts/hidalgocountytexas

*U.S. Census Bureau QuickFacts: Starr County, Texas*. (n.d.). U.S. Census Bureau. Retrieved September 7, 2022, from https://www.census.gov/quickfacts/starrcountytexas

*U.S. Census Bureau QuickFacts: Willacy County, Texas*. (n.d.). U.S. Census Bureau. Retrieved September 7, 2022, from https://www.census.gov/quickfacts/willacycountytexas

*US Census Bureau Quick Facts: Cameron County, Texas*. (n.d.). Census Bureau. Retrieved September 7, 2022, from https://www.census.gov/quickfacts/cameroncountytexas

Walker, W. E., Haasnoot, M., & Kwakkel, J. H. (2013). Adapt or perish: A review of planning approaches for adaptation under deep uncertainty. *Sustainability*, *5*(3), 955-979.

Walker, W. E., Rahman, S. A., & Cave, J. (2001). Adaptive policies, policy analysis, and policy-making. *European journal of operational Research*, *128*(2), 282-289.

*Willacy County, TX Cities & Towns*. (n.d.). MapSof.net. Retrieved September 7, 2022, from https://www.mapsof.net/willacy-county/cities

Xu, Z., Lachlan, K., Ellis, L., & Rainear, A. M. (2019). Understanding public opinion in different disaster stages: A case study of Hurricane Irma. *Internet Research*.

Zevenbergen, C., Rijke, J., van Herk, S., Chelleri, L., & Bloemen, P. J. T. M. (2016). Towards an adaptive, flood risk management strategy in The Netherlands: An overview of recent history. *International Journal of Water*, *3*, 140.

**Appendix A. Words Removed from Analysis**

'from','roma','subject','re', 'edu', 'use','cameron', 'county', 'starr', 'willacy', 'hidalgo', 'city', 'harlingen', 'page','update','jim', 'hogg', 'val', 'verde', 'brooks', 'dimmit', 'edwards', 'kenedy', 'kinney', 'maverick', 'webb', 'zapata', 'lower', 'rio', 'grande', 'valley', 'regional', 'flood', 'risk', 'event', 'hazard', 'management', 'mitigation', 'control', 'planning', 'action', 'plan', 'plans', 'group','going','think','region','well','view','okay','latest','version','area','thing','really','one', 'want','comment','see','transcript','dont','yeah','right','need','yes','said','thank','complete','sir','question','thank','time','help','maybe','something','board','sorry','yall','going','sign','approve', 'motion', 'carries','opposed','minutes','favor','entertain','exported','good','morning','sir','aye','aug','first','point','today','give','much','take','included','kristina','david','garza','mean','state','committee','jaime','megan','projects','standards','aye','work','things','anything','joe','hino','lot','way','areas','task','item','commissioner','look','ill','done','project','goal','last','kind','standard','process','say','year','able','two','include','part','three','word','chairman','discussion','move','aug','let', 'completed', 'rev', 'com', 'broadcast', 'starting', 'attendees', 'listen', 'mode', 'good', 'afternoon', 'everyone', 'web', 'begin', 'shortly', 'please', 'mute', 'microphones', 'next', 'slide', 'please', 'get', 'started', 'like', 'items', 'know', 'participate', 'todays', 'taken', 'screenshot', 'example', 'attend', 'interface', 'reference', 'currently', 'listening', 'using', 'computers', 'speaker', 'system', 'default', 'would', 'prefer', 'join', 'phone', 'please', 'select', 'phone', 'call', 'audio', 'pane', 'dial', 'information', 'displayed', 'please', 'enter', 'technical', 'questions', 'audio', 'access', 'issues', 'may', 'box', 'know', 'anyone', 'difficulty', 'joining', 'web', 'may', 'call', 'enter', 'phone', 'number', 'phone', 'information', 'public', 'chat', 'box', 'member', 'public', 'difficulty', 'connecting', 'web', 'may', 'call', 'meeting', 'using', 'audio', 'access', 'audio', 'phone', 'number', 'access', 'code', 'provided', 'please', 'note', 'phone', 'number', 'listening', 'purposes', 'additionally', 'please', 'type', 'questions', 'members', 'presenters', 'box', 'public', 'opportunity', 'provide', 'comments', 'ask', 'questions', 'designated', 'public', 'periods', 'noted', 'meeting', 'agenda', 'individuals', 'registered', 'using', 'online', 'public', 'registration', 'form', 'called', 'upon', 'muted', 'provide', 'comments', 'appropriate', 'filled', 'public', 'registration', 'form', 'member', 'public', 'wishing', 'provide', 'oral', 'public', 'comments', 'please', 'either', 'fill', 'form', 'raise', 'hand', 'designated', 'public', 'period', 'using', 'hand', 'button', 'provided', 'screen', 'next', 'slide', 'please', 'hello', 'welcome', 'everyone', 'name', 'james', 'bro', 'niko', 'ski', 'director', 'department', 'texas', 'water', 'development', 'next', 'slide', 'please', 'calling', 'afternoons', 'meeting', 'order', 'would', 'like', 'roll', 'call', 'voting', 'members', 'make', 'sure', 'establish', 'quorum', 'calling', 'meeting', 'order', 'list', 'call', 'name', 'please', 'answer', 'either'